\documentclass{article}
\usepackage[utf8]{inputenc}
\usepackage{authblk}
\usepackage{setspace}
\usepackage[margin=1.25in]{geometry}
\usepackage{graphicx}
\graphicspath{ {./Figures/} }
\usepackage{subcaption}
\usepackage{amsmath}
\usepackage[ruled]{algorithm2e}
\newcommand{\KPR}{kernels-per-row}
\newcommand{\KRN}{kernel-row-number}

\usepackage[style=nejm, 
citestyle=numeric-comp,
sorting=none]{biblatex}
\title{MaizeEar-SAM: Zero-Shot Maize Ear Phenotyping} %Kernel Phenotyping} %Trait Extraction for Plant Phenotyping}% for Maize Phenotyping}

\author[1]{Hossein Zaremehrjerdi}
\author[2,6]{Lisa Coffey}
\author[3]{Talukder Jubery}
\author[2]{Huyu Liu}
\author[4]{Jon Turkus}
\author[4]{Kyle Linders}
\author[4]{James C. Schnable}
\author[2,6]{Patrick S. Schnable\thanks{Corresponding author: pschnable@iastate.edu}}
\author[1,3,5,6]{Baskar Ganapathysubramanian\thanks{Corresponding author: bg@iastate.edu}}

\affil[1]{Department of Electrical and Computer Engineering, Iowa State University, Ames, IA, USA}
\affil[2]{Department of Agronomy, Iowa State University, Ames, IA, USA}
\affil[3]{Translational AI Research and Education Center, Iowa State University, Ames, IA, USA}
\affil[4]{Department of Agronomy, University of Nebraska-Lincoln, Lincoln, NE, USA}
\affil[5]{Department of Mechanical Engineering, Iowa State University, Ames, IA, USA}
\affil[6]{Plant Science Institute, Iowa State University, Ames, IA, USA}

\date{}

\begin{document}

\maketitle

%%%%%% Abstract %%%%%%
\begin{abstract}
Quantifying the variation in yield component traits of maize (Zea mays L.), which collectively determine the overall productivity of this globally significant crop, plays a critical role in plant genetics research, plant breeding, and the development of improved agronomic practices. Grain yield per acre is a function of the number of plants per acre $\times$ ears per plant $\times$ number of kernels per ear $\times$ average kernel weight. Kernels per ear is the product of the number of kernel rows per ear (\textit{\KRN}) and the number of kernels per row (\textit{\KPR}). Traditional manual approaches to quantifying these two traits, which exhibit differential responses to genetic and environmental factors, are time-consuming limiting the scale of data collect for either trait. Recent automation efforts using image processing and deep learning face challenges of high annotation costs and uncertain generalizability. We address these issues by exploring Large Vision Models for zero-shot, annotation-free maize kernel segmentation. Utilizing an open source large vision model -- Segment Anything Model (SAM) -- we segment individual kernels in RGB images of maize ears and implement a graph-based algorithm to calculate \KPR. Our approach effectively identifies \KPR~across diverse maize ears, demonstrating the potential of zero-shot learning using foundation vision models coupled with image processing routines to enhance automation and reduce subjectivity in agronomic data collection. All our codes are open-sourced to democratize these frugal phenotyping approaches. 
\end{abstract}

%%%%%% Main Text %%%%%%

\section{Introduction}

\label{sec:intro}
\subsection{Background and Motivation}

Maize is a major food and feed crop as well as an important industrial feedstock for ethanol production and an array of other industrial products. Given its widespread use and substantial contribution to the world economy \cite{usda2019projections}, there is significant interest in increasing maize yield per acre. Accurately estimating the yield of maize, and yield components, however, remains a complex process, heavily influenced by environmental conditions, soil fertility \cite{khanal2018integration}, pest pressures, and the crop's genetic makeup \cite{millet2019genomic}. Maize ear phenotyping plays a crucial role in assessing yield components. The total kernel number per ear is commonly considered a critical yield component, exhibiting high sensitivity to environmental conditions and management practices. This trait contributes more significantly to yield variation than kernel weight \cite{Sadras:2012tw,Shewry:1998uz,Ruiz:2022tz}. The total kernel number is determined by two factors: the number of \KPR~and the \KRN~\cite{wang2022impacts}. 

Kernel-row-number is strongly influenced by plant genetics and less so by environment. The determination of \KRN~occurs in a very short period after ear meristem initiation and is not substantially affected by the environment \cite{Abendroth:2011,strachan2004corn}.  In contrast, the number of \KPR~in maize is more susceptible to environmental effects than \KRN, responding to factors such as drought \cite{Wang:2019ug}, temperature \cite{Niu:2021,wang2022impacts}, and nitrogen availability \cite{Ruiz:2022tz} throughout its development from ear shoot initiation to grain fill \cite{Abendroth:2011,strachan2004corn}. This trait's development occurs in three main stages. During ovule development, genetic factors are influential, though less rigidly than for \KRN. At pollination, environmental stress can affect fertilization success. During grain fill, stress can lead to kernel abortion or death. %This process shows how kernels per row remain sensitive to environmental conditions, unlike the more stable kernel row number. 
While a well-developed ear of maize can bear over 650–800 kernels, unfavorable conditions can significantly reduce this number. Because \KRN~is consistent across most years and environments, the change in the number of \KPR~reflects the yield response to environmental changes. This relationship provides valuable insights into phenotypic plasticity, is crucial for breeding stable, high-yielding varieties and identifying stress-resilient genotypes. 

%Consequently, identifying genotypes with enhanced resilience to such stressors is vital for food security and economic stability. Hence, there is a pressing need for a dependable and efficient technique to extract these traits to assist agronomists in discovering high-yielding and more climate-resilient genotypes.

Traditionally, the process of counting \KPR~has been manual and labor-intensive and is susceptible to errors due to the subjective process of manual counting. The manual counting process involves several steps: first, typically a representative row is selected based on the average length and condition of the ear, deliberately avoiding extremes in row lengths and areas affected by disease or abnormalities. For ears with fewer distinct rows, the most well-defined row is often chosen in an effort to maintain consistency and prevent inaccuracies due to inadvertently switching from one row to another. Once a representative row is identified, kernels are counted from the base to the tip, excluding underdeveloped and non-viable kernels. An alternate approach to approximating the \KPR~involves determining the total number of kernels -- by automatically shelling the entire ear and then using a kernel counting machine to count the total number of kernels. This is preceded (before shelling) by manually counting the \KRN. By dividing the total kernel count by the number of rows, one can estimate the \KPR~trait. However, this approach is also noisy and tedious: there is up to 20\% inaccuracy in the machine counting process (see Fig~\ref{fig:kernel_counting_machine} in appendix), and manual row counting of \KRN~ is still required. Moreover, the large volume of samples—often exceeding 10,000 ears annually in research labs— frequently makes manual scoring of either \KPR~or \KRN~infeasible. %requires a careful balance between precision and efficiency in high-throughput phenotyping workflows. 
Unusual ear morphologies further complicate accurate counting. %In this work, we propose averaging counts from multiple rows per ear to improve accuracy.

%Despite efforts to standardize this measurement, variability persists due to the subjective nature of row selection and inherent differences among rows. 

Therefore, developing an approach to directly estimate the \KPR~would significantly streamline the phenotyping process. This automated pipeline could lower the resources required to quantify \KPR~(an important yield component trait) across a larger proportion of all maize experiments allowing researchers to better disambiguate and quantify the different genetic and environmental factors which can influence \KRN or \KPR. Furthermore, such automated pipelines will enable the widespread collection of new, more valuable and accurate trait data that are currently infeasible to collect.

%To mitigate subjectivity and enhance accuracy, averaging counts from multiple rows per ear has been considered. However, given that labs analyzing maize traits often process over 10,000 ears annually, efficiency remains a priority, influencing current methodologies in high-throughput phenotyping.
 
Recent advances in image processing, machine learning, and deep learning technologies have significantly transformed this field \cite{singh2025use}, offering more reliable methods for large-scale phenotyping. Nonetheless, the application of deep learning in agriculture, especially for estimating the number of \KPR, remains relatively under-explored. Previous attempts targeting similar traits, such as the total number of kernels \cite{wu2020kernelcounting}, have faced limitations, including the need for annotated images, specific lighting conditions, and the inability to handle variations in ear sizes, kernel colors, and the presence of multiple ears within a single image.

\subsection{Related Work}

% first work
Wu et al.\cite{wu2020kernelcounting} propose a method for estimating the total number of kernels on a maize ear using a single image. The process involves extracting features from the image, followed by the recognition of each kernel in the RGB images of the maize ear. The process employs a variety of algorithms, including the environmentally adaptive segmentation algorithm (EASA) for environmentally adaptive plant segmentation, a median filter and a Wallis filter for pre-processing, an improved Otsu method combined with a multi-threshold and row-by-row gradient-based method (RBGM) for kernel separation, and a segmentation method that integrates a genetic algorithm with an improved pulse coupled neural network. While the method demonstrates promising results, it has challenges. The accurate counting of kernels necessitates the identification and separation of adjacent kernels, a task that can be complex due to factors such as the narrow color gradient between kernels, the presence of simultaneous corner-to-corner, edge-to-corner, and edge-to-edge touching, and the irregular sizes and shapes of kernels on the same maize ear. Despite these challenges, the method offers a cost-effective and adaptable approach for kernel estimation, making it a valuable contribution to the field.
% end of first work

% second work
Khaki et al.\cite{khaki2020corndetection} employed a sliding window approach for kernel detection, where a Convolutional Neural Network (CNN) classifier assesses each window for the presence of a kernel. Post classification, a non-maximum suppression algorithm is applied to eliminate redundant and overlapping detections. Finally,  windows identified as containing a kernel are passed to a regression model, which predicts the (x, y) coordinates of the kernel centers. Grift et al. \cite{grift2017kernelcounting} propose a method to handle imperfect segmentation, where some kernel areas are connected to two or more other kernel areas. When a bridged kernel is detected based on area discrimination, the image is traced back to the original grey-scale image, to allow applying the de-bridging algorithm. The de-bridging algorithm consists of extraction of the bridged area from the kernel map, retrieving the grey scale bridged area from the original ear image, and segmentation of the grey scale bridged area using Otsu's method with a local threshold.
  
% end third work

% summary of the limitations of the above methods that will be addressed by this method.
%Previous methods in maize ear phenotyping are typically tailored for specific backgrounds or lighting conditions and trained on limited datasets, making them not generalizable across different environments or crop varieties. The semi-automated method for counting maize kernels has several limitations. It relies on manual counting, which can introduce human error. It may struggle with accurately identifying individual kernels in certain situations. The method's effectiveness with ears affected by disease, mold, or field conditions is unclear. Lastly, the method has demonstrated a potential error range in kernel counting, indicating possible inaccuracies. Additionally, these deep learning methods require substantial amounts of annotated data, which is both time-consuming and labor-intensive to collect. Given that we are dealing with new genotypes, encountering unseen kernel shapes in the input data is also challenging. Furthermore, no objective definition for the kernel row structure could be mathematically formulated. Manual counting of kernels per row is a tedious task, begging for automation to enhance efficiency and accuracy.

As described above, existing maize ear phenotyping methods, while valuable, face certain challenges. Wu et al.'s approach encounters difficulties with kernel separation due to color similarities and shape irregularities. Khaki et al.'s sliding window technique, though effective, may have limitations with varying kernel sizes and image resolution. Grift et al.'s method addresses some segmentation issues but involves complex image processing steps. These methods often work best under specific conditions, for example, lighting and background, which can limit their broader applicability. They typically require large annotated datasets and may not perform optimally with unfamiliar kernel shapes from new genotypes. Most importantly, the lack of a standardized definition for \KPR~also presents a challenge for automation. These factors suggest a need for more versatile approaches to maize ear phenotyping that can maintain accuracy across different conditions and varieties.
\subsection{Our Contribution}

We develop an end-to-end automated pipeline that counts \KPR~on maize ears without relying on annotated data. This pipeline addresses the significant resource and time constraints associated with annotating data required for traditional deep-learning models. By incorporating Zero-Shot Learning (ZSL) and advanced segmentation models, our approach effectively handles new genotypes and unseen variations in kernel appearance, shape, or colors, as well as image resolution, enhancing robustness and generalizability. This ensures that our pipeline remains effective under diverse and unpredictable environmental conditions. Additionally, we employ graph theory to introduce a clear, mathematical definition to address the subjectivity traditionally associated with defining kernel row structures. This significantly improves the consistency and reliability of the data collection, facilitating more accurate genetic analyses and breeding decisions. Together, these advancements streamline the maize phenotyping process, significantly improving its efficiency, scalability, and precision. 
We provide our workflow as open-source code to democratize the utilization of such frugal, high-throughput phenotyping approaches to the broader scientific community.

\section{Materials and Methods}

\label{sec:methods}

\subsection{Data Collection}
\label{subsec:dp}

\subsubsection{Dataset from the High-Intensity Phenotyping Sites (HIPS) Project}
\begin{figure}[!ht]
    \centering
    \begin{subfigure}{0.45\linewidth}
      \includegraphics[width=\linewidth]{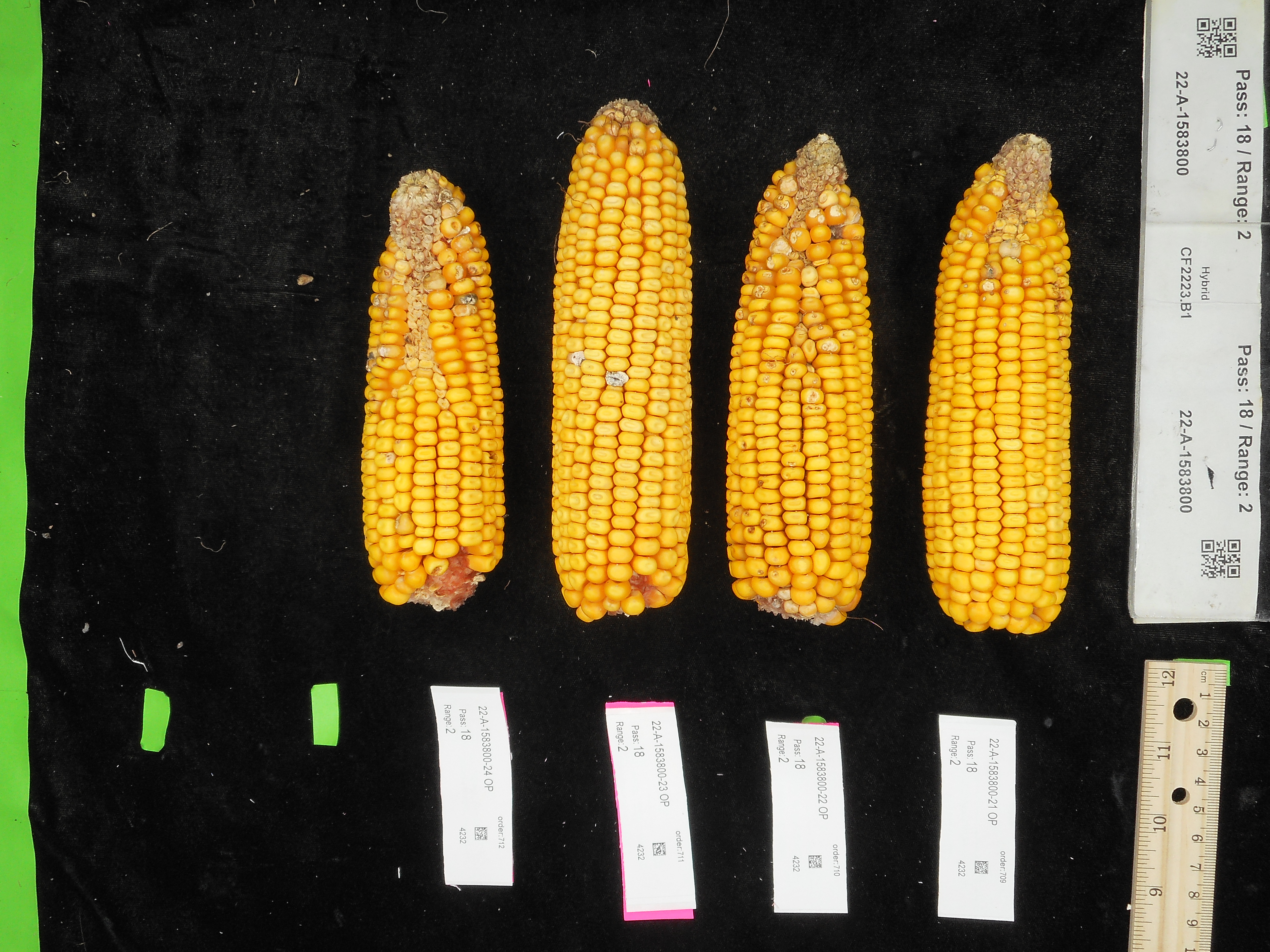}
    \end{subfigure}
    \hspace{5mm}
    \begin{subfigure}{0.45\linewidth}
      \includegraphics[width=\linewidth]{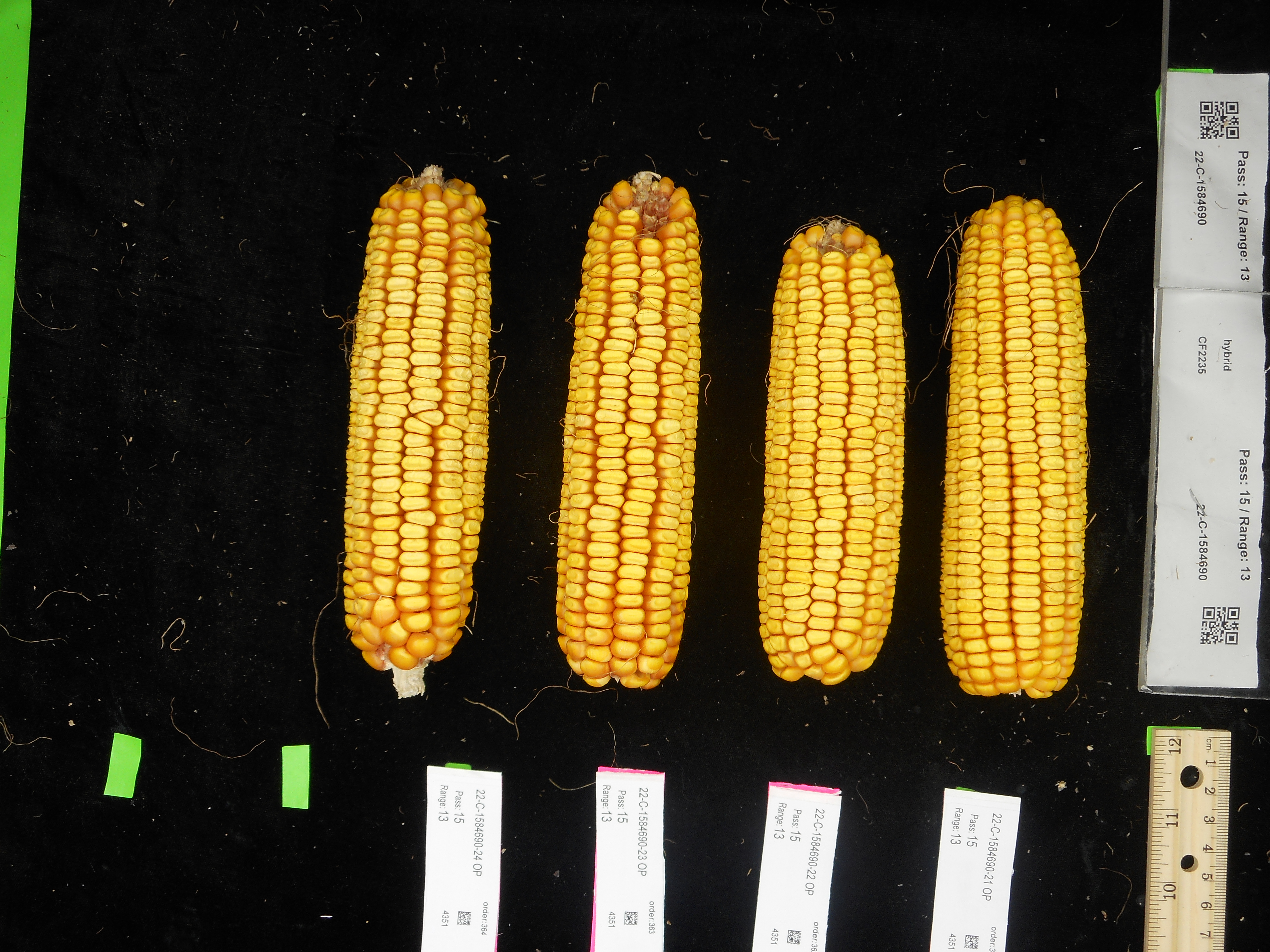}
    \end{subfigure}
  \caption{These images depict a sample image from the HIPS dataset. It includes 4 hybrid ears, each labeled with its ID beneath it. In the top right corner, there are QR codes corresponding to each tray of images. All ears in each tray are associated with the same genotype.}
    \label{fig:Hips_example}
\end{figure}

The input dataset comes from a USDA-funded HIPS project, which includes 84 hybrid maize lines grown at six locations across Iowa (IA) and Nebraska (NE). Open-pollinated ears were manually harvested from each research plot in the fall, just before machine harvesting commenced across all fields. The hand-harvested ears were then transported to data collection facilities for manual trait analysis. Within a single plot, ears were photographed together using a Nikon Coolpix S3700 digital camera positioned about 18 inches above an imaging frame, directly over a downward-facing ring light. Figure \ref{fig:Hips_example} shows a sample image captured by our imaging pipeline.

Each ear was laid out on a black backdrop, oriented so the tips of the ears are all pointed to the top (as shown in Figure \ref{fig:Hips_example}) The field row tag, displaying the row plot ID, was placed on the side of the frame and secured with a plexiglass sheet. Tags identifying individual ears were placed adjacent to each ear. For spatial reference, a ruler was positioned in the corner of the frame in 2022, which was replaced by a color grid in 2023. The layout was designed to include negative space around each ear to facilitate automated segmentation of individual ears. Once the placement of all components was finalized, colored tape was used on the backdrop to ensure consistent positioning in all 1,000 images captured. A specific camera zoom depth was selected to accommodate larger ears within the frame while minimizing unused pixel space. This zoom setting was used consistently  across all images. The number of harvested ears per row varied across the hybrid fields. In total, approximately 1,000 images of hybrid ears were taken, covering over 4,000 ears.
\subsubsection{Ground Truthing}
\begin{figure}[h]
    \centering
    {\includegraphics[width=0.7\linewidth]{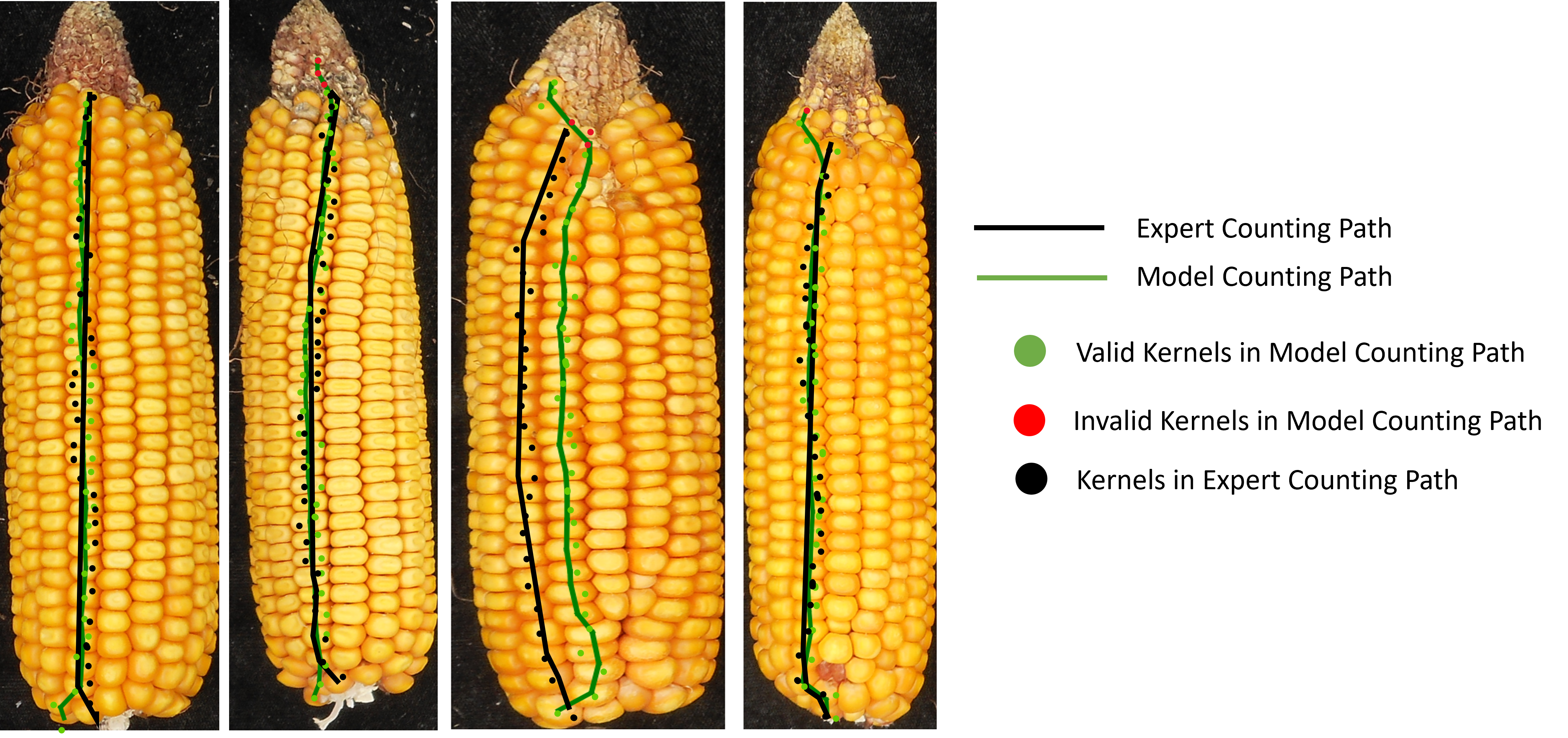}}
    
    \caption{This image illustrates two distinct path types: the green path, generated by our automated pipeline, and the black path, generated by a human expert. Additionally, red dots indicate invalid kernels deemed immature or unhealthy by the expert, green dots represent valid kernels that should be counted by the model, and black dots denote kernels counted by the expert.} 
    
    \label{fig:groun_truthig}
\end{figure}

% mention that subset of an image selected and selection method for ground truthing. 
To ensure that our evaluation captures the full diversity of our dataset, we asked experts to select approximately 150 ears for detailed manual phenotyping and baselining. This subset is representative of the entire dataset, allowing us to comprehensively assess our model's performance. Domain experts annotated the dataset in two ways to ensure comprehensive ground truth verification. The first task involved traditional practices for counting \KPR. Experts selected the most representative row visible in the 2D image, marked by black lines in Figure.~\ref{fig:groun_truthig}, which was typical of average length and free from isolated damage, disease, or poor development. They then drew a line across the image to delineate the path of the row and placed dots on each fully developed kernel along this path. 

The second task required experts to annotate based on a path selected by our automated pipeline. In this approach, experts marked and counted kernels along the model-selected path, as shown in Figure \ref{fig:groun_truthig} by the green line, including any underdeveloped kernels that were included and noting any fully developed kernels that were overlooked by the model. They also observed and recorded any unusual kernel morphologies that might affect the accuracy of the kernel count or the selection of a representative row.

We conducted a test where the domain expert re-annotated the same images one month later without access to their previous annotations to quantify the subjectivity or repeatability of manual annotations. This approach ensured independence from earlier work and allowed us to assess human annotation consistency (subjectivity) over time.

\subsection{Workflow for automated extraction of the Kernel-Per-Row trait}

\begin{figure}
    \centering  \includegraphics[width=1\textwidth]{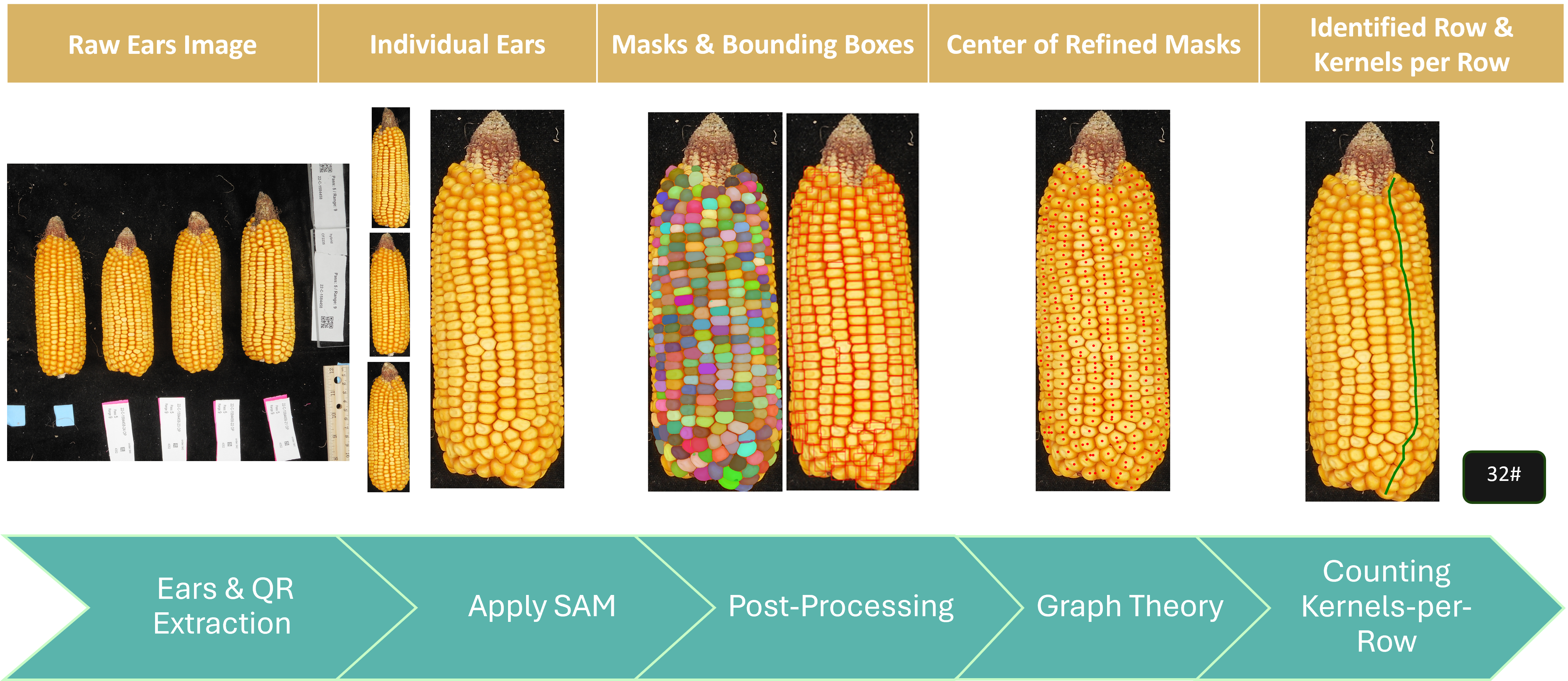}
    \caption{This figure demonstrates our approach to counting the \KPR~on maize ears, beginning with raw images that typically contain four to six ears. The first step involves extracting individual  ears and reading the QR codes from each tray, resulting in images of separated ears for further processing. Next, we apply the Segment Anything Model (SAM) to these isolated ears to generate masks and bounding boxes for each kernel. Following this, in our post-processing step, we refine these masks and compute the center points of each kernel. Finally, utilizing graph theory, we frame our analysis as a graph problem to identify the most representative row and count the kernels, focusing on fully developed kernels to ascertain the number of \KPR.}
    \label{fig:workflow}
\end{figure}

Our methodology for extracting \KPR~ for individual maize ears from image sets is executed through a series of steps, as shown in Figure \ref{fig:workflow}. These steps are Ears and QR code extraction, applying Segment-Anything-Model (SAM) and post-processing, graph theory, and \KPR~counting; we elaborate each step in the next sub-sections. Each image in our dataset typically features four to six maize ears that were designated as a raw ears image. The first step, which is ears and QR code extraction, involves extracting  ears out of images to prepare them for a more detailed examination. We detail this step in Section.~\ref{subsec:extraction_qr}. Once ears have been isolated, we apply the Segment-Anything-Model (SAM) for image segmentation, see Section.~\ref{subsec:sam}. Applying the SAM step allows us to distinguish and segment the individual kernels from each ear. Following segmentation, our post-processing phase (detailed in Section.~\ref{subsec:postprocessing}) involves performing quality control on each kernel's masks. This refinement step is crucial for ensuring the accuracy of subsequent analyses. After kernels have been effectively segmented, we create bounding boxes for each kernel. The center points of these bounding boxes represent the center of each kernel.

We employ graph theory (see Section.~\ref{subsec:graph}) and construct an adjacency matrix using these center points. This matrix is used for calculating the shortest path across the maize ear, from the first kernel to the last. Our study's primary trait of interest is the number of nodes along this shortest path, or, in other words, the number of \KPR. In a final refinement step, we filter out any immature kernels identified on this path, ensuring that our analysis focuses solely on fully developed kernels.

\subsubsection{Extracting Maize Ears and QR Codes: Image Analysis Techniques}
\label{subsec:extraction_qr}

To extract each maize ear from an image, we first convert the image to the HSV color space to enhance color-based segmentation. We then apply a binary threshold to the Value (V) channel using OpenCV, which helps distinguish the maize ears from the background. Using OpenCV functions, we find the contours of the maize ears in the thresholded image and automatically trace these contours to allow for their individual extraction. Subsequently, we employ the Pyzbar library to read QR codes from each extracted maize ear image.The QR code data, containing each maize ear's genotype, location, and agronomic treatment details, is integrated as metadata. This metadata can be linked to trait values for downstream analyses, including Genome-Wide Association Studies (GWAS).

\subsubsection{Applying the Segment Anything Model for Kernel Masking}
\label{subsec:sam}

Zero-shot learning is a powerful machine learning paradigm in which the model can handle unseen data that it has not been explicitly trained on. This approach contrasts sharply with supervised learning, where the model is trained on labeled data specific to the tasks it will perform. An example of this innovative approach is the Segment Anything Model \cite{kirillov_segment_2023}, which is a versatile and robust foundation model (for segmentation) due to its training on nearly 1 billion masks. The model's ability to generate output masks for various images without prior exposure to similar tasks showcases its proficiency in zero-shot learning. The model employs a Vision Transformer (ViT) \cite{dosovitskiy_image_2021} structure, which supports its ability to handle new, unseen images effectively in a zero-shot manner.

The Segment Anything Model (SAM) is tailored for image segmentation tasks, and is composed of three interconnected components. The first component, the image encoder, processes raw image data into a compact form known as image embedding, capturing essential information about the image. The second component, the prompt encoder, transforms prompts, which can be any information indicating what to segment in an image, into a processable form. The final component, the mask decoder, combines the embeddings from the image encoder and the prompt encoder to predict a segmentation mask, representing the segment of the image that corresponds to the prompt. This structure allows the SAM to respond appropriately to any prompt at inference time, making it a powerful tool for real-time, interactive image segmentation tasks.

%In practice, we apply this model's capabilities in two steps: First, we use SAM to segment individual ears from raw images. Following this, 

The \textit{Automatic Mask Generator}, a feature within the model, is tasked with creating precise masks for all kernels in an individual ear. 
The \textit{SamAutomaticMaskGenerator} generates segmentation masks for an entire image. It begins by creating a grid of point prompts over the image. The model processes these points in batches, generating masks for each batch. Given that our image contains many objects, we increased the number of generated points to allow the model to extract all possible objects in the image. These masks are then filtered based on their predicted quality and stability. The process is repeated for different image segments to ensure comprehensive coverage. After generating masks for all segments, the class removes duplicates and any small disconnected regions or holes in the masks. Finally, the masks are encoded in a chosen format (binary mask, uncompressed Run-Length Encoding (RLE), or COCO RLE) and returned with additional information such as the bounding box, area, predicted quality, and stability score.
Given the complexity and high number of kernels typically present, we systematically varied and identified robust choices for parameters. In particular, the number of generated points (\textit{points\_per\_side}) was set to 80 for robust performance.

\subsubsection{Enhancing Mask Quality in Post-Processing}
\label{subsec:postprocessing}
Upon extracting the corresponding masks, we perform post-processing to ensure each mask accurately corresponds to a single kernel. This is crucial because some masks generated during the previous stage may represent multiple kernels or include areas not relevant to any kernel. The post-processing includes checking (a) the area of each mask to ensure it should be less than an upper bound (here, 10,000 pixels) and larger than a lower bound (here, 1000 pixels), and (b) the model's certainty is above a set threshold (here, set to 0.93). Additionally, we compute the Intersection over Union (IoU) for the masks, as described in Equation \ref{eq:iou}. If the IoU exceeds 0.4, we discard the larger mask as it encompasses multiple kernels.
\begin{equation} \label{eq:iou}
\text{IoU} = \frac{\text{Area of Intersection}}{\text{Area of Union}}
\end{equation}

Once we refine the model's masks by selecting the appropriate ones and removing any over-extended masks, we calculate the center of each kernel, utilizing the bounding boxes associated with them.

\begin{algorithm}[h!]
    \small
    \caption{Finding the Neighborhood to Make the Adjacency Matrix}
    \label{algo:neighborhood_adj_matrix}
    \SetKwInOut{Input}{Input}
    \SetKwInOut{Output}{Output}
    \Input{List of kernel centers $\mathbf{V} = \{v_0, v_1, \ldots, v_N\}$}
    \Output{Adjacency matrix $\mathbf{A}$ of size $N \times N$}
    \textbf{Initialize:}\\
    \For{each $v_i \in \mathbf{V}$}{
        Compute distances $D_{ij}$ to all other $v_j \in \mathbf{V}$\;
        Sort $D_{ij}$ and select the five nearest neighbors $\mathbf{N}_i$ for $v_i$\;
    }
    \textbf{Refine Neighbors:}\\
    \For{each $v_i \in \mathbf{V}$}{
        \For{each pair $(v_j, v_k)$ in $\mathbf{N}_i$}{
            Compute angle $\theta_{ijk}$ between $v_i\to v_j$ and $v_i\to v_k$\;
            \If{$\theta_{ijk} < 20^\circ$}{
                Remove the farthest neighbor from $\mathbf{N}_i$\;
            }
        }
    }
        \textbf{Construct Adjacency Matrix:}\\
    \For{each $v_i \in \mathbf{V}$}{
        \For{each $v_j \in \mathbf{N}_i$}{
            Set $\mathbf{A}_{ij} = \text{Distance}(v_i, v_j)$\;
            $\mathbf{A}_{ji} = \mathbf{A}_{ij}$ \tcc{Ensure the matrix is symmetric}\;
        }
    }
    \Return{$\mathbf{A}$}
\end{algorithm}

\subsubsection{Graph Theory and Kernel Counting and Filtering Immature Kernels}
\label{subsec:graph}
Having obtained the center coordinates of all kernels, we treat each kernel center as a node in a graph, creating an adjacency matrix with entries denoting the Euclidean distance between centers of kernels based on their image (i.e., pixel) locations. We apply Dijkstra's algorithm to navigate the graph of a maize ear, calculating the shortest path from the second bottom-most to the second topmost kernel. This approach enables us to accurately determine the number of \KPR. Based on our extensive numerical experiments, using the second bottom-most and second topmost kernels as start and end nodes produced the most reliable result. This strategy helps avoid errors due to detached kernels or non-kernel elements that are sometimes present at the ear ends.

Beyond distance, we incorporate additional measures to accurately identify kernel neighbors, recognizing that distance alone may not fully depict the true spatial relationships between kernels. When two kernels are aligned in the same row close to the target kernel, relying solely on distance could lead to inaccurate neighbor classification. To address this, as outlined in Algorithm \ref{algo:neighborhood_adj_matrix}, we initially identify the five nearest neighbors based on the distance to the target kernel. We then evaluate the angles formed with neighboring kernels; if the (vertical) angle is less than 20 degrees -- we chose this value empirically based on extensive observations -- we conclude that the two kernels are aligned in a row. Thus, we eliminate the other kernels from consideration in the adjacency matrix. This refinement step ensures that our adjacency matrix accurately reflects the physical arrangement of kernels on the ear, which is crucial for the success of Dijkstra's algorithm in our \KPR~counting task. Finally, to filter out immature kernels, we start from the top of the path and proceed downward, evaluating if each kernel exceeds a specific size threshold (indicating a mature kernel, here, 2000 pixels). We subtract the number of these immature kernels from the total kernel count per row that is calculated.

Dijkstra's algorithm determines the shortest path between nodes in a graph \cite{cormen2022introduction}. The algorithm functions by "visiting" each node, starting from a predetermined initial node, and keeping a record of the shortest known distance to each node from the starting point. For each new node it visits, it updates the recorded distances to the neighboring nodes if it finds a shorter path. It repeats this process, always visiting the next unvisited node with the shortest recorded distance until it has found the shortest path to all nodes. Here, we apply this algorithm to find the shortest path from the second bottom-most to the second topmost kernel, thereby providing a number of \KPR~in a maize ear.

\subsection{Multi-Path Strategy for Precise Kernel Distribution Analysis}
\begin{figure}[h!]
    \centering  \includegraphics[width=0.45\textwidth]{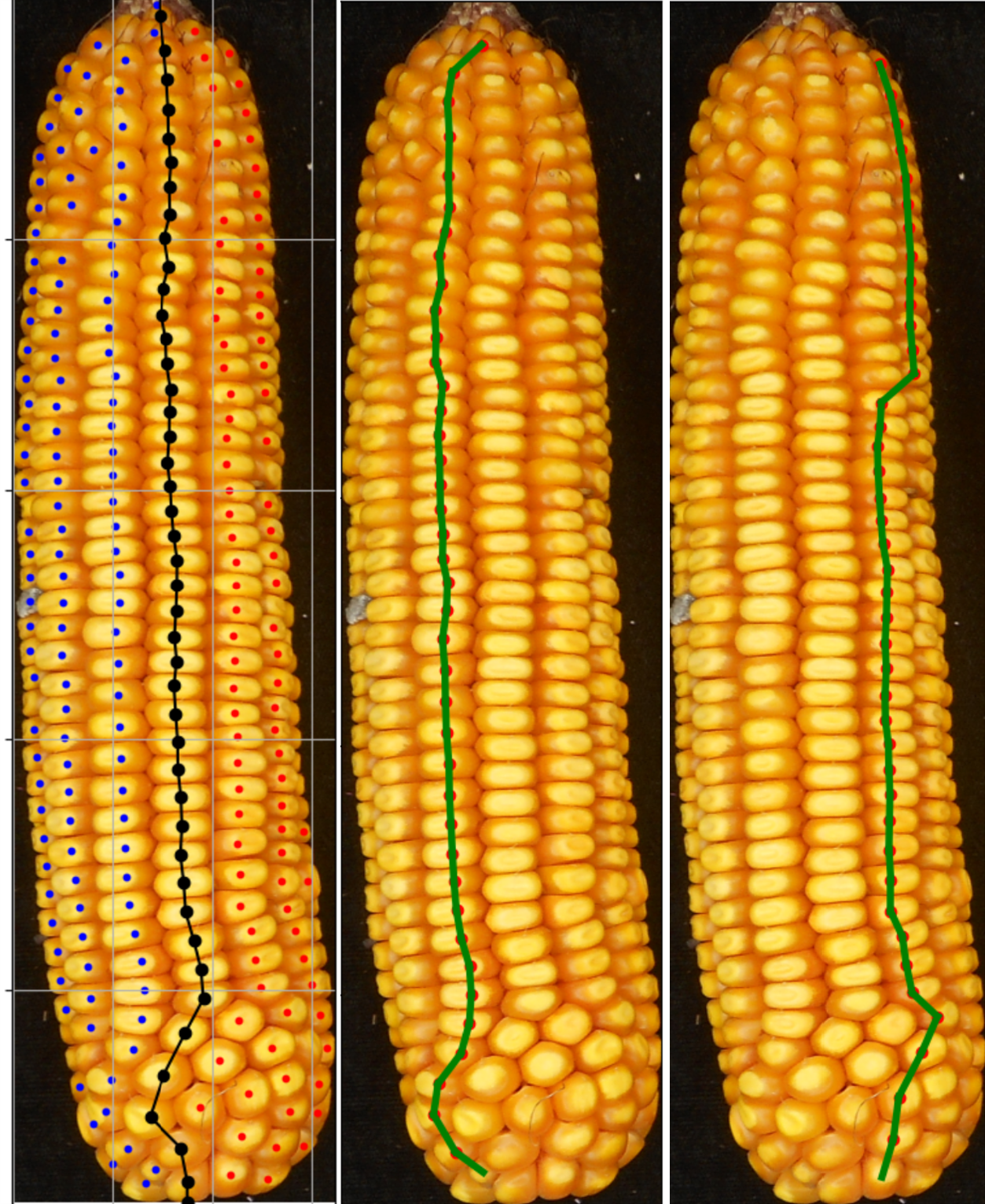}
    \caption{Three paths: In the left image, the black dotted  line indicates the central path that splits the ear into two halves. The middle and right images respectively illustrate the detailed paths of the left and right halves, as derived from the initial segmentation}
    \label{fig:threePath}
\end{figure}

To enhance the robustness of our methodology, we generate and analyze three distinct paths for each ear rather than relying on a single path. This multi-path approach addresses potential irregularities in kernel placement, particularly at the top of the cob, where immature kernels can lead to significant asymmetry, thus ensuring more statistically consistent outcomes. Initially, we identify a central path, as described in section \ref{subsec:graph}, which forms the basis for generating the additional paths on either side. This central path bisects the ear into two halves, creating left and right segments. For each segment, we further determine the shortest path, culminating in a total of three paths for detailed examination. This strategy, illustrated in Figure \ref{fig:threePath} and detailed in Algorithm \ref{algo:compute_path_distribute_kernels}, enhances the reliability of our \KPR~counting by averaging the results across these paths, thereby mitigating the effects of anomalies or errors that might occur with a single-path analysis.

%In  Algorithm \ref{algo:compute_path_distribute_kernels}, a boundary is created by linking consecutive kernels along the central path, extending to the image's edges. Kernels are then classified into left or right arrays based on their x-coordinates relative to this boundary. Finally, a path-finding function is applied on each of these arrays to identify the shortest paths. The rationale behind this approach is to ensure a comprehensive assessment of the ear's structure, accounting for potential irregularities in kernel placement that might skew results if only a single path were considered. By evaluating multiple sections of the ear and averaging these observations, we achieve a more accurate and representative measure of the number of kernels per row.  

\begin{algorithm}[h!]
    \small
    \caption{Compute Path and Split Kernels Based on Position}
    \label{algo:compute_path_distribute_kernels}
    \SetKwInOut{Input}{Input}
    \SetKwInOut{Output}{Output}
    \Input{~List of kernels in the maize, image dimensions $(\text{max\_width}, \text{max\_height})$}
    \Output{~Kernel classification into left and right arrays based on a central path}
    
    \textbf{Compute Central Path:}\\
    \For{each pair of consecutive kernels on the middle path}{
        \text{Locate the middle path kernel and connect them with a line}\\
        \text{For the first and last kernel, connect to a point with $y = 0$ and $y = \text{max\_height}$ with the same $x$}\\
    }
    
    \textbf{Classify Kernels Based on Central Path:}\\
    \For{each kernel in the maize not on the middle line}{
        \text{Find the corresponding point on the polyline that has the same $y$}\\
        \eIf{kernel's $x >$ point's $x$ on the line}{
            \text{Add kernel to the right kernels array}\\
        }{
            \text{Add kernel to the left kernels array}\\
        }
    }
    
    \textbf{Apply Path Finding:}\\
    \text{Choose either the right or left kernels array}\\
    \text{Apply the find path function on the chosen array}\\
    
    \Return{Left and right kernels arrays, central path}
\end{algorithm}

\section{Results}
\label{sec:results}

In this section, we evaluate the performance of our automated approach (MaizeEar-SAM) against human expert annotation, focusing on two distinct types of errors. Additionally, we will discuss the infrastructure used for running the code and the associated time consumption.

\subsection{Evaluation of the Model for Counting Kernel-per-Row}

The first type of error pertains to the accuracy of MaizeEar-SAM in counting \KPR, irrespective of path selection. To evaluate this, the human domain expert who created the ground truth examined the paths generated by our model. They then counted the kernels along these specified paths to establish a benchmark for comparison. This process allows us to gauge MaizeEar-SAM's counting \KPR~accuracy without considering the path selection criteria.

The second type of error assesses MaizeEar-SAM's ability to select paths and count the \KPR. We have a human expert choose a path and then count the \KPR~along that path. Our analysis followed two procedures. First, we compared the \KPR~counts from MaizeEAR-SAM-selected path to those where humans manually selected the path and counted the kernels on that path, to evaluate the model's accuracy in path selection. Next, we created three paths using MaizeEAR-SAM and compared the average kernel count from these three paths to the ground truth count from the human-selected path. This study provides insights into the MaizeEar-SAM's path selection and kernel counting capabilities.

Before reporting results, we clarify our evaluation metrics: We calculate the ratio of the expected value of the predicted number of \KPR~to the ground truth (actual) number of \KPR. To express this as a percentage, we multiply the ratio by 100. This percentage represents how close the expected predictions are to the human counted values.

\begin{figure}[h!]
    \centering
    \begin{subfigure}{0.48\linewidth}
      \includegraphics[width=\linewidth]{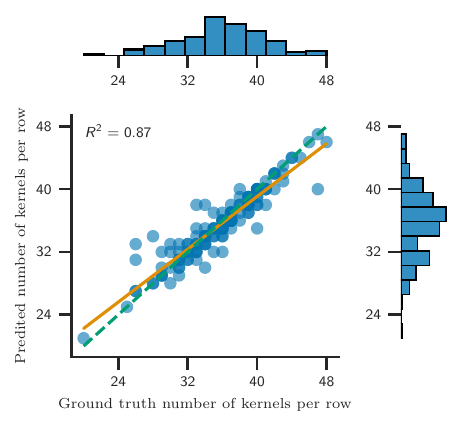}
      \caption{Predicted kernel count in MaizeEar-SAM path vs. ground truth kernel count in the same path, featuring 136 data points}
    \end{subfigure}
    \hspace{1mm}
    \begin{subfigure}{0.48\linewidth}
      \includegraphics[width=\linewidth]{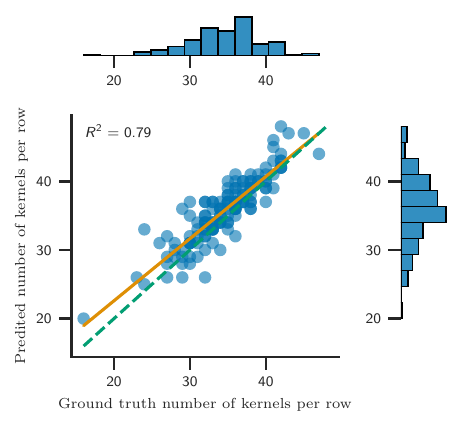}
      \caption{Predicted kernel count in MaizeEar-SAM path vs. ground truth kernel count in the expert path, featuring 136 data points}
    \end{subfigure}
    \caption{Comparative analysis of path selection and kernel counting. (a) This plot demonstrates MaizeEar-SAM's performance in counting kernels, including maturity-based kernel filtering. This is compared with a scenario where MaizeEar-SAM selects the path, but an expert performs the kernel counting, allowing for a nuanced evaluation of model accuracy versus expert judgment. (b) This plot contrasts the outcomes where, on one side, the path is selected by  MaizeEar-SAM and counted by an expert, and on the other side, both path selection and counting are conducted by an expert, underscoring the differences in subjectivity between automated and manual approaches.}
    \label{fig:model_comparisons1}
\end{figure}

\begin{figure}[h!]
    \centering
    {\includegraphics[width=3in]{    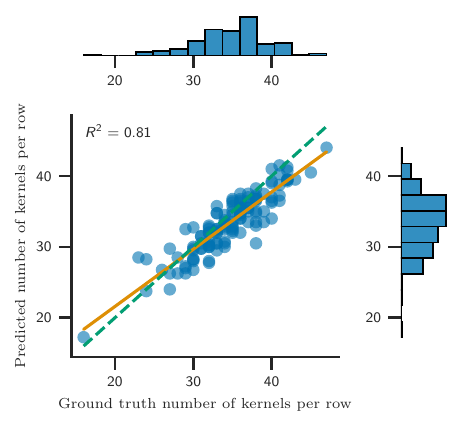}}
    \hspace{5mm} % Space between the images
    
    \caption{Average predicted kernel count from three different paths generated by MaizeEar-SAM, compared to ground truth kernel count in the expert path, featuring 130 data points}
    \label{fig:three_path_comparison}
\end{figure}

Figures \ref{fig:model_comparisons1} and \ref{fig:three_path_comparison} offer an in-depth analysis of MaizeEar-SAM's performance in path selection and counting \KPR, set against the ground truth which is generated by human expertise. Figure \ref{fig:model_comparisons1} consists of two comparisons. Part (a) evaluates  MaizeEar-SAM's proficiency in automatically selecting paths and counting \KPR. This plot shows the assessment of MaizeEar-SAM's \KPR~counting accuracy, regardless of the path selection process. In part (b), the comparison extends to scenarios where the model selects a path that is then counted by an expert versus instances where an expert undertakes both tasks. This comparison seeks to reveal the accuracy and selection criteria differences between automated processes and manual efforts.

Figure \ref{fig:three_path_comparison} focuses on contrasting MaizeEar-SAM's automated path generation and counting \KPR~capabilities with the manual approaches used by experts. In this analysis, MaizeEar-SAM generates three distinct paths and performs counting \KPR~for each path, then compares the average of those three paths to the counts obtained when an expert selects the path and performs the counting. This comparison illustrates how using multiple paths and averaging those paths could improve the process of counting the \KPR. This highlights the key challenges and factors in developing automated solutions for \KPR, comparing the efficiency and reliability of automated versus manual path selection and counting.

\subsection{Deployment Performance and Time Analysis}
When deploying our framework on a large dataset, it is crucial to have an accurate estimate of the execution time. The table below breaks down the time for each part of MaizeEar-SAM, allowing us to identify the most time-consuming sections. The computations were performed on a machine with an A100 GPU with 80GB RAM. To record the time for each step of the algorithm, the pipeline was executed on the entire dataset, and we included the histogram of the distribution of the number of \KPR~across this dataset in Figure.~\ref{fig:NKR_histogram}. Note that phenotyping of each maize ear takes less than 20 seconds. Thus, this approach of automated phenotyping can extract maize ear traits for 4,000 ears in one day. Note that this does not account for the time required to photograph the ears. 
\begin{table}[h!]
\centering
\begin{tabular}{|c|c|c|}
\hline
Extracting(sec) & Mask Generate(sec) & Row Counting(sec) \\
\hline
3.5  & 14.4 & 0.9  \\
\hline
\end{tabular}
\caption{Time metrics for model performance. All times are per maize ear.}
\label{tab:time_table}
\end{table}

\section{Discussion}
\label{sec:conclusion}
\begin{figure}[!]
    \centering
    {\includegraphics[width=3in]{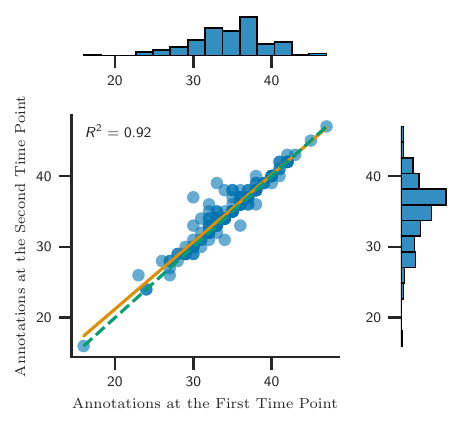}}
    \caption{Variability in kernel count was assessed by the same person at two different time points, based on 136 data points.}
    \label{fig:subjectivity_plot}
\end{figure}

In this study, we leveraged the vision foundation model's zero-shot capabilities to automate counting \KPR. Initially, each individual ear was extracted from the original multi-ear image. The Segment Anything Model (SAM) was applied to each ear to mask each kernel. After refining these kernels, graph theory was used to calculate the shortest path from bottom to top. The kernels along this path were counted and reported as the number of \KPR. Statistical robustness is ensured by averaging the number of \KPR~by considering three distinct rows. This approach produces reproducible results. Moreover, the automation of this process not only expedites previously labor-intensive annotations but also could effectively replace the tedious job of manually counting kernels, thereby freeing up valuable human resources for more strategic tasks. Finally, having a mathematical definition of this trait removes the subjectivity (inter-rater and intra-rater) associated with a human expert performing this count.

Figure.~\ref{fig:subjectivity_plot} illustrates this point. We requested a human expert to annotate the same set of images but at two distinct times (approximately one month apart). The human expert spent several hours annotating 136 individual maize ears. Figure.~\ref{fig:subjectivity_plot} shows the (intra-rater) subjectivity with their count of number of \KPR~only matching with an $R^2$ of 0.92. This suggests that an individual counts kernels on the same ear in different ways at different times, introducing a degree of subjectivity into the data. \footnote{This raises an interesting point: How much of the variability between the human expert and our automated measurements is contributed by human expert subjectivity? There is no clear way to resolve this question, which further makes the case for a reproducible, objective approach to defining and automatically measuring this trait.} This variability can impact the power of Genome-Wide Association Studies (GWAS), as the noise from subjective counting practices reduces the statistical power of GWAS to detect associations between genetic markers and trait variation. This variation highlights the need for an objective approach to defining traits of interest. Because this maize trait is sensitive to environmental changes, such as drought, identifying the genes that control the stability across environments of this trait paves the way for developing more robust maize hybrids, benefiting farmers and food security. 

Future research will focus on deploying this approach to a large dataset of collected maize ear images exhibiting a diverse range of kernel and ear, followed by genetic analysis (GWAS), and breeding decision support. Additional improvements include updated workflows for identifying optimal maize path types, such as distinguishing when a twisted path might better represent a row on the ears (see Fig.~\ref{fig:non_standard_maize} in the appendix for a non-standard scenario). Additionally, we are extending this approach to extract additional traits like the number of rows, the total kernel count, and characteristics of the kernel alignment along rows.

\section*{Acknowledgments}
\subsection*{Funding}
This material is based upon work supported by the AI Research Institutes program supported by NSF and USDA-NIFA under AI Institute: for Resilient Agriculture, Award No. 2021-67021-35329, Agriculture and Food Research Initiative grant no. 2021-67021-35329/project accession no. IOWW-2021-07266, grant no. 2020-68013-30934/project accession no. IOW05612, grant no. 2020-67021-31528/project accession no. IOWW-2020-01463 and grant no. 2018-67021-27845/project accession no. IOW05533 and by the National Science Foundation under Grant No. CNS-2125484. 

\subsection*{Data Availability}

All codes are available for use at https://bitbucket.org/baskargroup/maizeear-sam 
\printbibliography

\newpage
\appendix

\section{Additional Results}
\label{sec:sample:more-example}
Additional representative results are shown below. 
\begin{table}[h!]
\centering
\begin{tabular}{|c|c|c|}
\hline
Test Image index & Real kernel number  & predicted kernel number \\
\hline
1  & 38 & 39  \\
\hline
2  & 32 & 31  \\
\hline
3  & 40 & 40  \\
\hline
\end{tabular}
\label{tab:my_label}
\end{table}

\begin{figure}[h!]
    \centering
    \begin{subfigure}{0.22\linewidth}
      \includegraphics[height=6cm]{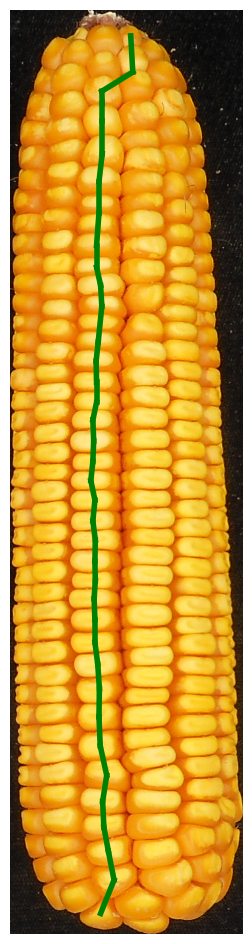}
      \caption{Image 1}
    \end{subfigure}
    \hspace{2mm}
    \begin{subfigure}{0.22\linewidth}
      \includegraphics[height=6cm]{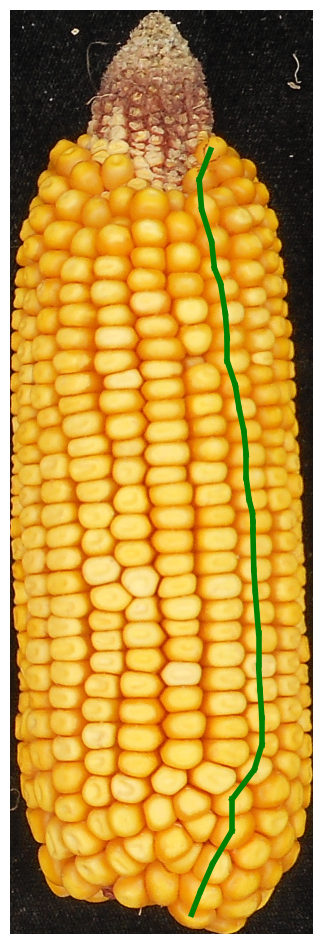}
      \caption{Image 2}
    \end{subfigure}
    \hspace{2mm}
    \begin{subfigure}{0.22\linewidth}
      \includegraphics[height=6cm]{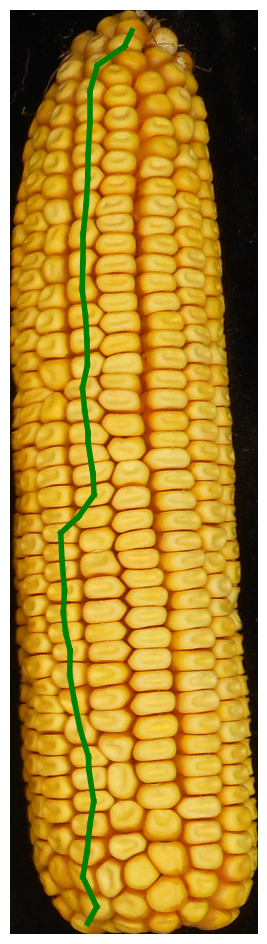}
      \caption{Image 3}
    \end{subfigure}
    \caption{Examples comparing real and predicted kernel counts.}
    \label{fig:model_comparisons}
\end{figure}

\begin{figure}[h!]
    \centering
    {\includegraphics[width=0.65\linewidth]{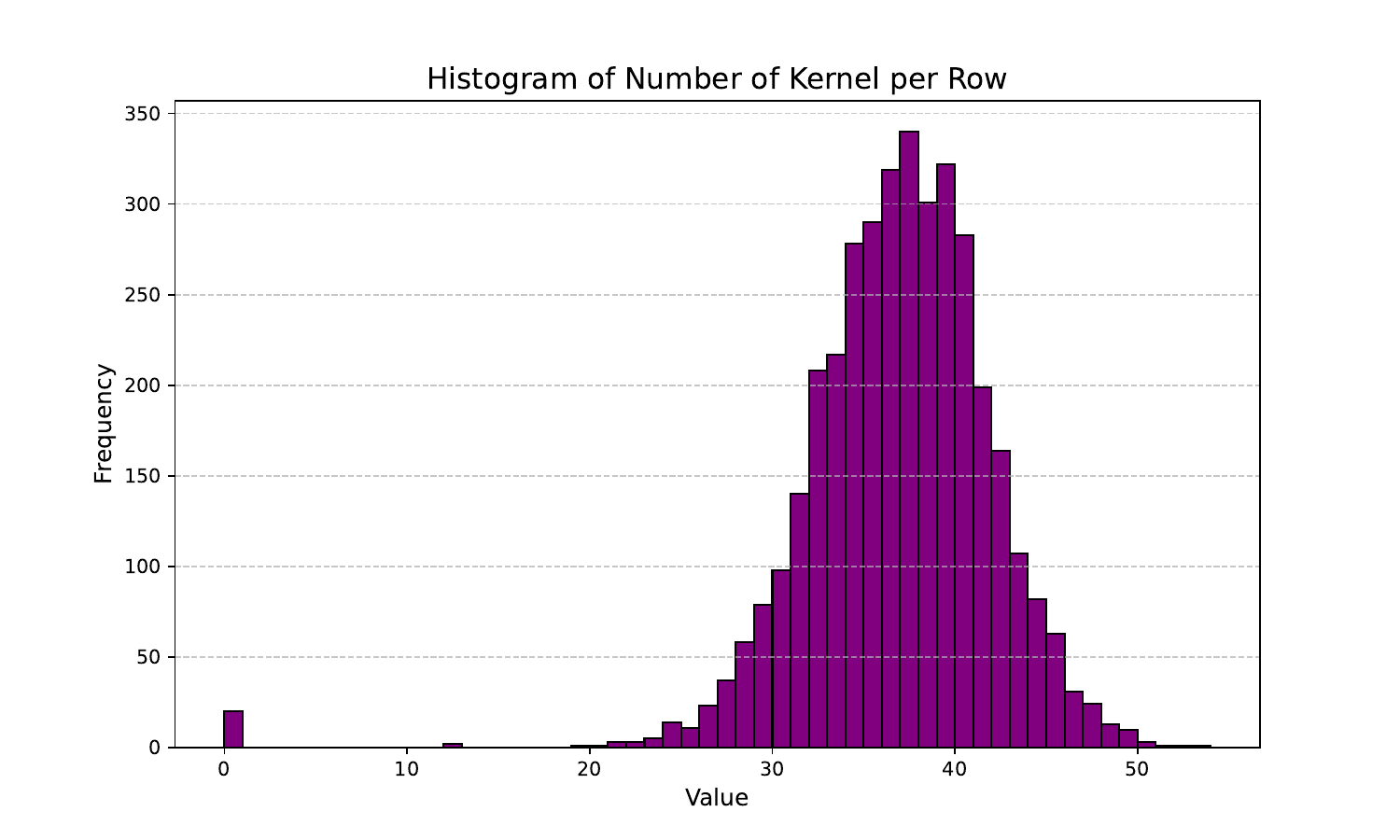}}
    \caption{Histogram of the Number of Kernels-per-Row across 4000 ears}
    \label{fig:NKR_histogram}
\end{figure}

We ran MaizeEar-SAM on all images and plotted the histogram for the entire dataset. Approximately 4,000 maize samples were used to generate the histogram shown in Figure.~\ref{fig:NKR_histogram}.

\begin{figure}[!]
    \centering
    {\includegraphics[width=0.75\linewidth]{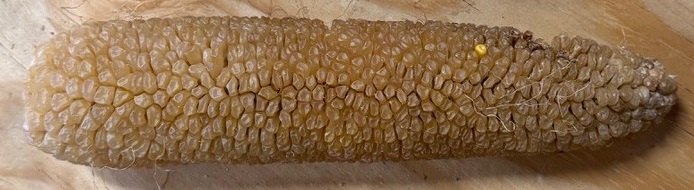}}
    \caption{Maize ears without a clear kernel arrangement, making identification of rows difficult}
    \label{fig:non_standard_maize}
\end{figure}

Figure.~\ref{fig:non_standard_maize} illustrates examples where manual counting of \KPR~can become subjective. These non-standard cases emphasize the need for an objective definition of \KPR~which MaizeEAR-SAM provides.

\begin{figure}[!]
    \centering
    {\includegraphics[width=0.5\linewidth]{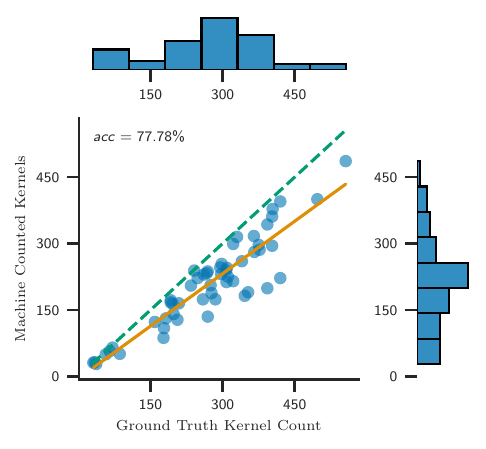}}
    \caption{The plot illustrates the inaccuracy in estimating the total number of kernels, where the x-axis represents the (ground truth) human count, and the y-axis shows the machine counted values}
    \label{fig:kernel_counting_machine}
\end{figure}

Figure.~\ref{fig:kernel_counting_machine} illustrates the potential issues of automatically shelling the entire ear and then using a kernel-counting machine to count the total number of kernels. The figure compares human counting of all the kernels in a shelled ear versus a machine automatically counting the kernels. We observe a non-trivial amount of variability (up to 20\% inaccuracy) in the machine counting process.

\end{document}